\documentclass[preprint]{vgtc}               %

\graphicspath{{figures/}{pictures/}{images/}{./}} %

\usepackage{times}                     %

\usepackage[usenames, dvipsnames]{color} %
\usepackage{enumitem}
\usepackage{wrapfig}

\usepackage{mathptmx}                  %

\usepackage[font=footnotesize]{subfig}

\usepackage{graphicx}

\newif\ifauthornotes
\authornotestrue  %

\newcommand*\samethanks[1][\value{footnote}]{\footnotemark[#1]}

\onlineid{0}

\vgtccategory{Research}

\vgtcinsertpkg

\title{Opening the Black Box of 3D Reconstruction Error Analysis with VECTOR}

\author{Racquel Fygenson\thanks{Kazi Jawad and Racquel Fygenson contributed equally to this work.}  \thanks{e-mail: fygenson.r@northeastern.edu}\\ %
        \scriptsize Northeastern University %
\and Kazi Jawad\samethanks[1]  \thanks{e-mail: kjawad@wetafx.co.nz}\\ %
    \scriptsize W\=et\=a FX %
\and Isabel Li\\ %
     \parbox{0.9in}{\scriptsize \centering Art Center \\ College of Design}
\and Francois Ayoub\\ %
     \parbox{1.6in}{\scriptsize \centering Jet Propulsion Laboratory \\ California Institute of Technology}
\\
\and Robert G. Deen\\ %
     \parbox{1.6in}{\scriptsize \centering Jet Propulsion Laboratory \\ California Institute of Technology}
\and Scott Davidoff\\ %
     \parbox{1.6in}{\scriptsize \centering Jet Propulsion Laboratory \\ California Institute of Technology}
\and Dominik Moritz\\ %
     \parbox{1.1in}{\scriptsize \centering Carnegie Mellon University}
\and Mauricio Hess-Flores\\ %
     \parbox{1.6in}{\scriptsize \centering Jet Propulsion Laboratory \\ California Institute of Technology}
}

\abstract{
Reconstruction of 3D scenes from 2D images is a technical challenge that impacts domains from Earth and planetary sciences and space exploration to augmented and virtual reality. Typically, reconstruction algorithms first identify common features across images and then minimize reconstruction errors after estimating the shape of the terrain. This \textit{bundle adjustment} (BA) step optimizes around a single, simplifying scalar value that obfuscates many possible causes of reconstruction errors (e.g., initial estimate of the position and orientation of the camera, lighting conditions, ease of feature detection in the terrain). Reconstruction errors can lead to inaccurate scientific inferences or endanger a spacecraft exploring a remote environment.
To address this challenge, we present VECTOR, a visual analysis tool that improves error inspection for stereo reconstruction BA. VECTOR provides analysts with previously unavailable visibility into feature locations, camera pose, and computed 3D points. 
VECTOR was developed in partnership with the Perseverance Mars Rover and Ingenuity Mars Helicopter terrain reconstruction team at the NASA Jet Propulsion Laboratory. We report on how this tool was used to debug and improve terrain reconstruction for the Mars 2020 mission.

} %

\keywords{Computer vision, stereo image processing, optimization, error analysis, uncertainty, SLAM, SfM, robotics.}

\nocopyrightspace

\begin{document}

\firstsection{Introduction}

\maketitle
\label{sec:introduction}
Stereo image processing to reconstruct a three-dimensional scene from two-dimensional images is an increasingly important capability in many domains. Such algorithms brings new analytic perspectives to fields as broad and diverse as geosciences~\cite{James:2012:JGR}, cultural heritage~\cite{Guidi:2004:TIP}, archaeology~\cite{magnani2020digital,dellepiane:2013:JCH}, Earth~\cite{dewitt2000elements} and planetary science, and robotics for space exploration~\cite{grisetti2010tutorial,Matthies:2007:IJCV,maimone:2007:JFR}. As a camera or set of cameras moves around a scene, perceived motion in the plane of the camera(s) (i.e., \textit{parallax}) can be used to determine the distance of an object. As the same objects are viewed across multiple images, a powerful family of algorithms that includes
simultaneous localization and mapping (SLAM)~\cite{mur:2015:Robotics} and structure-from-motion (SfM)~\cite{agarwal:2009:ICCV} can be used to build a 3D model of the scene.

These stereo reconstruction algorithms generally identify and match visual features, or \textbf{tiepoints}, corresponding to points in the scene (i.e., \textbf{tracks}) across all available images, and estimate a 3D point cloud via \textbf{triangulation}, given known or estimated camera positions and orientations (i.e., \textbf{pose}). These algorithms then run \textit{bundle adjustment}~\cite{Triggs00}~(\textbf{BA}) to minimize \textit{total reprojection error} across many parameters-- often thousands of cameras and 3D points~\cite{levenberg:1944:QAM,lour09, Agarwal:Ceres:2022}--and generate a single goodness-of-fit statistic. This single scalar value---which represents the complex set of interacting parameters in stereo reconstruction---provides no information about how individual residuals, which can be influenced by inaccurate pose and structure estimates, impact the optimization process. This metric also provides no visibility into how particular images, lighting conditions, camera positions, or details of the morphology of the remote environment might interact to create inaccuracies in a particular output. The impact of these unknowns compounds in domains where high accuracy terrain reconstruction is critical to outcomes, like science or space exploration where there is no ground truth and inaccurate reconstruction can lead to false results or risking billion-dollar spacecraft. Alarmingly, while there has been an explosion of stereo algorithms across many fields, there has been little work visualizing errors of such processes and the effects of individual outliers on the accuracy of results.

In response, we developed VECTOR, a visual analytics application for the \textbf{V}isual \textbf{E}diting of \textbf{C}amera \textbf{T}iepoints, \textbf{O}rientations and \textbf{R}esiduals. VECTOR is an interactive visualization tool that shows the sensitivity of scene structure before BA (i.e., during feature tracking, pose estimation, and triangulation), as well as after BA (i.e., the interaction between all parameters), better characterizing model output, errors, and noise. Furthermore, VECTOR users can analyze outputs visually, which speeds up the debugging of algorithms. Thus, VECTOR accelerates algorithmic development and enables more informed operational decisions.

This paper presents 
VECTOR and its application to characterizing stereo reconstruction errors in mission operations for the Mars Perseverance Rover and Ingenuity Helicopter. In particular, this paper makes the following contributions:
\begin{enumerate}[topsep=4pt,itemsep=-1ex,partopsep=1ex,parsep=1ex]
    \item Introduces the first visual analytics application to facilitate error analysis for stereo reconstruction algorithms.
    \item Describes the process of error analysis for stereo reconstruction in the context of planetary robotic exploration.
    \item Illustrates how visual encoding and interaction can address the challenges of terrain reconstruction.
\end{enumerate}

\begin{figure*}[t]
 \centering
 \includegraphics[width=\textwidth]{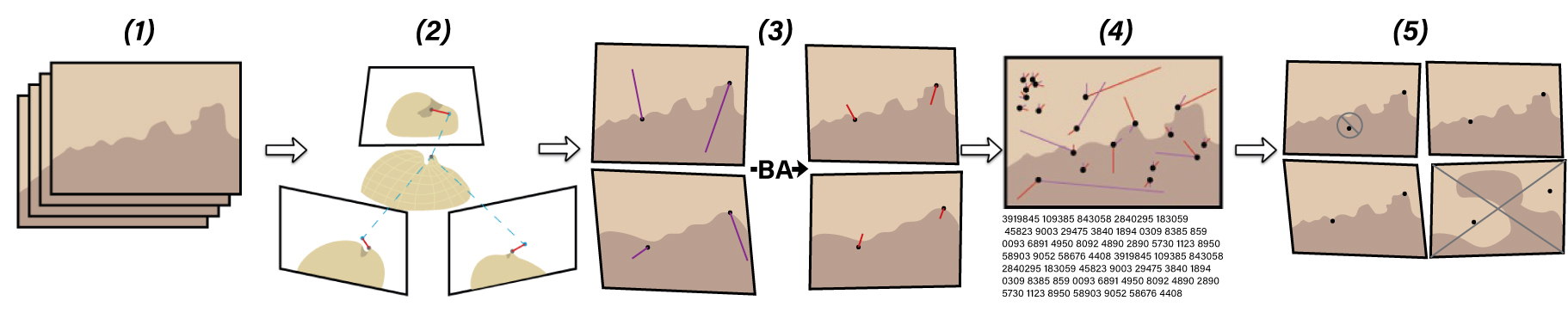}
 \vspace{-2mm}
 \caption{The stereo scene reconstruction process. Starting with a set of 2D images for stereo reconstruction (step 1), the scene reconstruction algorithm computes tiepoints and tracks, triangulates their 3D position and back-projects them into the original images creating residuals (step 2). Then, the BA algorithm optimizes camera poses and 3D structure to minimize residuals (step 3), outputting optimized residuals in each image along with descriptive statistics in XML (step 4). Analysts then manually remove erroneous tracks (e.g., top-left of step 5) and camera poses (e.g., bottom-right of step 5), and re-run the process.}
 \label{fig:algo}
 \vspace{-5mm}
\end{figure*}

\section{Background}
\label{sec:background}

In this section, we situate stereo reconstruction error analysis within the broader context of error analysis and uncertainty visualization in machine learning and robotics. We then unpack the challenges facing stereo reconstruction error analysis.

\subsection{Understanding Machine Learning Output}

VECTOR is motivated by a long history of using visualization and visual analytics to understand the performance of machine learning (ML) models. Recent work has focused on interactive explanation of model performance~\cite{hohman:2019:gamut}, or generation of  rules~\cite{wing:2019:rulematrix} or counterfactuals~\cite{wexler:2019:what-if} across broad classes of models from deep learning networks~\cite{hohman:2018:visual} to random forests~\cite{zhao:2019:iforest}, and in domains from bias detection~\cite{cabrera:2019:fairvis} to space exploration~\cite{Bae:2020:worldviews} and the physical sciences~\cite{wright:2023:ishmap}.

Within machine learning, this paper focuses on the computer vision problem of stereo reconstruction, which has broad applicability across domains looking to recreate remote locations digitally~\cite{James:2012:JGR,Guidi:2004:TIP,magnani2020digital,dellepiane:2013:JCH,dewitt2000elements,grisetti2010tutorial,Matthies:2007:IJCV,maimone:2007:JFR}. Research into error analysis of stereo reconstruction often focuses on mathematical approaches~\cite{Hartley04,Brooks01,Cheong98,Zhao96}, including ground-truth-based camera pose estimations~\cite{Rodehorst08}, geometric error extraction free of simplifying assumptions~\cite{Knoblauch_Factorization}, or volumetric explanations for the total reprojection error in BA~\cite{Recker12}. VECTOR builds on these mathematical approaches, serving as a visual aid to error analysis that allows users to unpack model outputs to pursue root causes of errors.

\subsection{Error Analysis for Stereo Reconstruction Algorithms}

Stereo systems for rover navigation typically receive as input a set of images (\cref{fig:algo} step 1), then compute features and tracks across them (\cref{fig:algo} step 2), and use linear algebra and optimization techniques to produce a map of the viewed scene as well as fine-tune the pose of the camera(s) mounted on the rover~\cite{maimone:2007:JFR,Matthies:2007:IJCV} (\cref{fig:algo} step 3) received via telemetry. Typically, reconstruction algorithms such as SfM and SLAM minimize total reprojection error of all computed points across all cameras via BA~\cite{Triggs00}.

The accuracy of a multi-view reconstruction relies on accurate feature tracking and matching~\cite{Lowe04,Bay06}. Tiepoints can be computed using dense or sparse algorithms~\cite{Lowe04,Bay06,Beauchemin95} and linked via tracks. Because errors accumulate in downstream computation, track matching accuracy strongly influences overall scene reconstruction fidelity~\cite{Hartley04}. Even robust estimation procedures~\cite{Hartley04} are sensitive to lighting conditions and occlusions.
Due to these errors, computed 3D points do not reproject exactly onto their initial tiepoint positions in each image (see \cref{fig:algo}, step 2). Hence, the objective of BA is to adjust parameters in such a way that the total reprojection error of the 3D points with respect to their corresponding tiepoints in each camera is minimized. This error corresponds to the sum of squares of \textbf{residual errors} for each 3D point reprojected onto each image plane with respect to its corresponding tiepoint.
This minimization can be achieved using algorithms such as \emph{Levenberg-Marquardt} (LM)~\cite{Lourakis04}.

In the absence of ground-truth information, BA is the only valid geometrical evaluation of accuracy, despite its tendency to miring in local minima.
Furthermore, BA outputs a single, minimized scalar, revealing little to nothing about the optimization process, which variables most affected its outcome, outliers that heavily skewed the process, and other valuable information which could be used to improve the accuracy of the reconstruction. For instance, if a small number of feature tracks with high residual error represent a large percentage of the total reprojection error, these tracks could be removed or corrected to allow BA to converge much faster, and with a greater probability of reaching the global minimum.

\begin{figure*}%
\centering
\includegraphics[width = 0.93\textwidth]{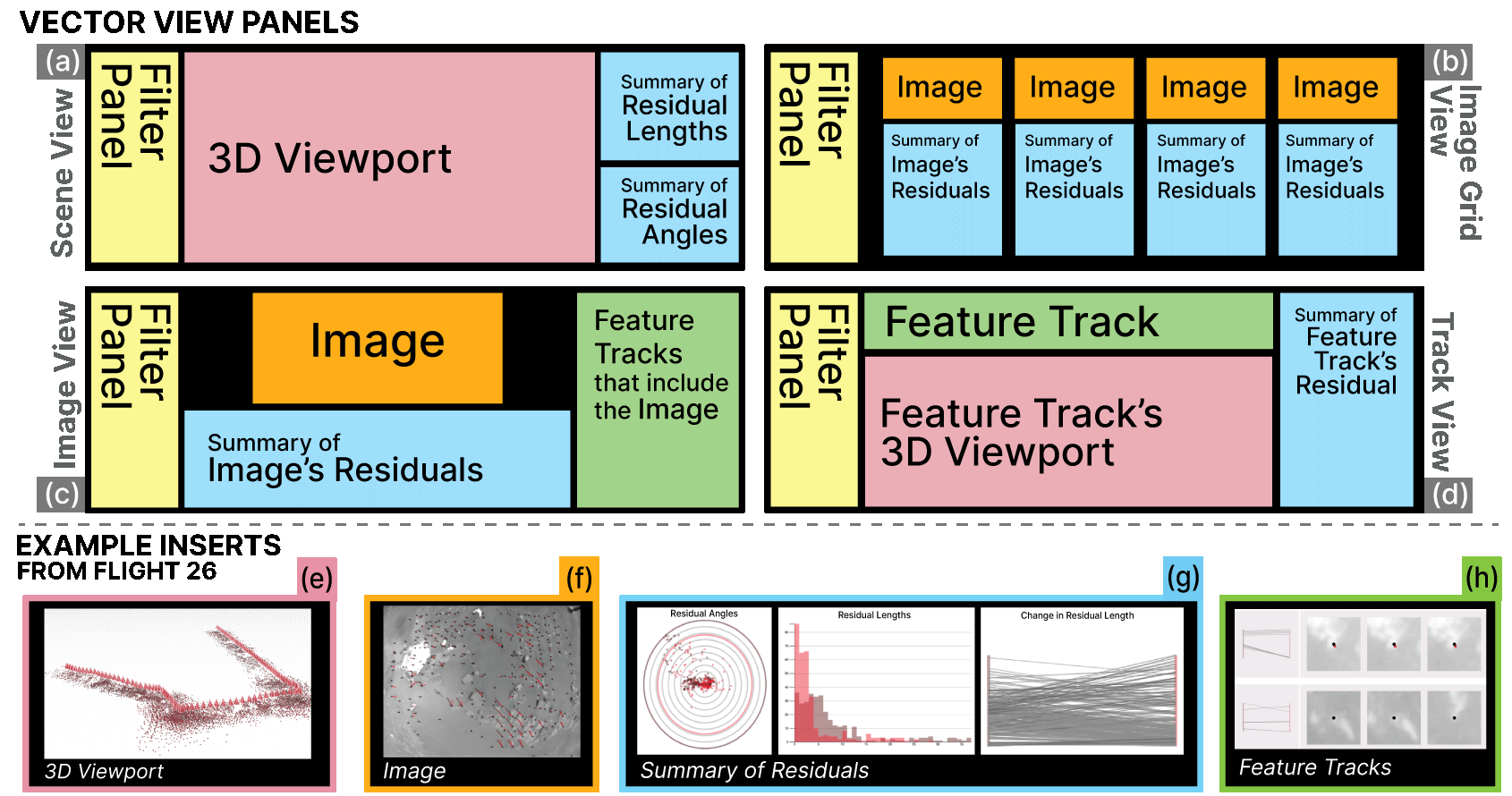}
\caption{Top: VECTOR panels which are used in tandem to detect and eliminate erroneous feature tracks and camera poses that adversely affect BA in stereo reconstruction. Bottom: example visualizations populating the shown panels.} 
\label{fig:flight26}
\vspace{-5mm}
\end{figure*}

\section{How Users Find Errors with VECTOR}

To understand how stereo reconstruction scientists approach error analysis, over one year, we partnered with a team of experts at the NASA Jet Propulsion Laboratory. We conducted semi-structured interviews, and created prototypes that we iteratively co-designed and evaluated~\cite{hendrie:2022:infovis}. Below, we detail the main user goals we uncovered, followed by how VECTOR's design facilitates each goal.

\label{sec:methodology}
\subsection{User Goals}
\label{sec:user_goals}

The variance in residual magnitudes and directions across feature tracks and adjusted cameras' location and orientation can reveal important information about parameters that is more useful than total reprojection error and can be used to improve BA performance. Ideally, residuals should be of minimum length and evenly distributed in 2-dimensional orientation. Stereo scientists evaluate the performance of BAs by analyzing patterns in the obtained residuals, which can number in the thousands. Resolving residuals is supported by three main goals:
\begin{enumerate}[topsep=4pt,itemsep=-1ex,partopsep=1ex,parsep=1ex]
    \item \textit{Identify specific feature tracks, or portions of these tracks, that contribute to large BA errors.}
    Feature tracks identify objects or terrain across multiple images, and typical reconstructions contain thousands of tracks. High residuals in even a few feature tracks can skew precise modeling of entire objects or scenes, so stereo scientists often attempt to locate and remedy tracks with large residuals.

    \item \textit{Delete high-error feature tracks.}
    After identifying feature tracks with large residuals or a concentration of residuals pointing in one direction, scientists can opt to remove these tracks and re-run BA. Feature tracks are characterized via unique identification strings. 
    
    \item \textit{Compare pre- and post- feature track removal models to evaluate improvement.}
    After finding and deleting error-prone feature tracks, scientists re-run BAs and investigate if the track deletion improved model performance. Prior to VECTOR, this was performed by toggling between long text files of each model's residuals--an intractable task for large datasets. 
\end{enumerate}

VECTOR has been key in understanding sources and magnitude of errors in SLAM, stereo, navigation and terrain-mapping methods at JPL, specifically for the Mars 2020 Perseverance rover and Ingenuity helicopter. VECTOR provides a cohesive platform for data filtering, editing, and reiterative analysis to address all three main user goals, adding more value to user workflows than equivalent non-interactive visualizations.
In the following subsections, we describe each of VECTOR's panels and how they were used to discover and correct issues in Ingenuity \textit{Flight 26}. This dataset consists of 122 images, where 11380 tracks were computed using a modified SIFT algorithm~\cite{Lowe04}. This dataset consists of straight linear motions of the helicopter as well as sharp turns, where triangulation is more likely to fail. In these areas, JPL scientists found that many ground points were computed incorrectly. Using VECTOR, operations teams removed outlier tracks and re-computed \textit{Flight 26}'s 3D reconstruction in a matter of minutes, instead of sinking resources into parsing a large text file manually, as would have been done previously. This VECTOR use case was conducted on a MacBook Pro laptop using the Google Chrome web browser.

\subsection{Data}

The data for SLAM and stereo reconstruction at JPL typically consists of XML files. For a given dataset, one XML file contains camera pose parameters and another contains feature tracking pixel values, XYZ 3D positions, and other relevant metadata. These files can reach a few GB in size and are stored on clusters. Manually parsing such large files is slow and makes it virtually impossible to spot relationships between spatial variables and pinpoint error causes. With VECTOR, a user can import these XML files directly and see them visualized in 2D and 3D space, visually compare data from different BAs, and find and resolve problematic residuals.
\subsection{VECTOR Panels}
VECTOR consists of four views which highlight information about different parts of the reconstruction process. Each panel supports different pattern finding and visual investigation that are integral to the stereo operators' reiterative process. VECTOR's panels are designed with Shneiderman's Visual Information Seeking Mantra in mind: overview first, zoom and filter, then details-on-demand~\cite{Shneiderman96}.

\subsubsection{Scene View}
\label{sec:scene-view}
\begin{figure*}[t!]
    \centering
    \includegraphics[width=0.8\linewidth]{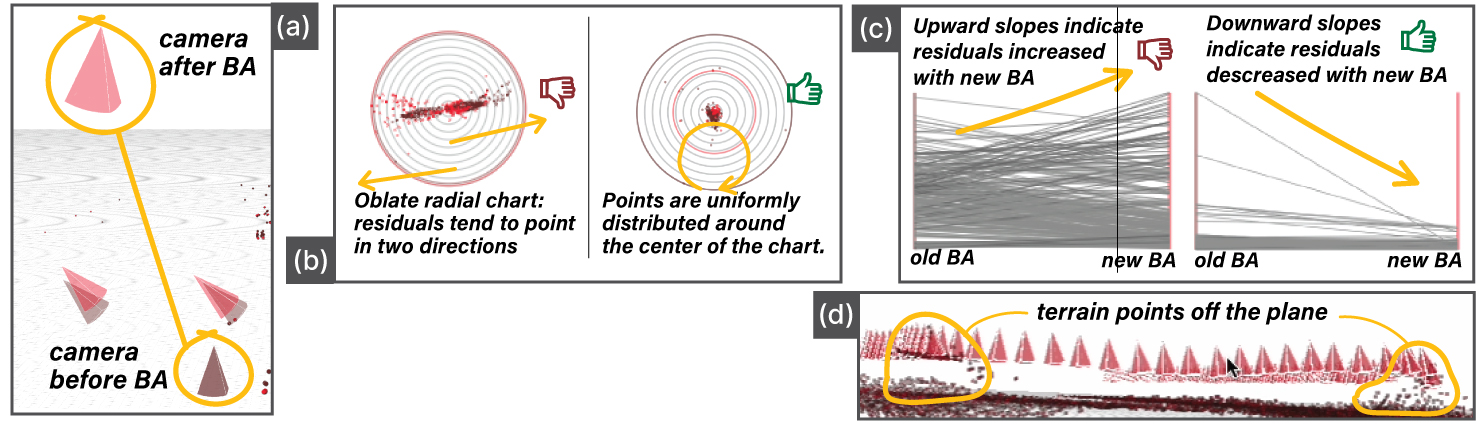}
    \caption{Annotated example visualizations from \textit{Flight 26}. (a) An example camera (circled) does not align with others. It could be removed to improve BA accuracy. (b) Ideally, residuals are short and uniformly distributed in angle. (c) Ideally, residuals decrease when a new BA is applied. (d) Inaccurate computed terrain points which should lie on the plane. }
    \label{fig:annotated}
    \vspace{-5mm}
\end{figure*}

VECTOR's entrypoint is the \textbf{\textit{Scene View}} (\cref{fig:flight26}a) which is a global representation of residuals across all tracks and cameras in the dataset. This view is broken down into an interactive BA viewport and a set of charts. The 3D viewport (\cref{fig:flight26}e) represents bundle-adjusted camera frustums (conic shapes) and feature track points (dots), with a two-tone color palette that highlights whether elements are pre- or post- BA. Users can navigate the viewport using simple click-and-drag mouse movements as well as keyboard controls to pan, zoom, roll, and tilt. 
Users can also click on individual cameras or scene points to navigate to their respective \textbf{\textit{Track Views}} or \textbf{\textit{Image Views}} for further detail. 
If a scene point lies noticeably far from others (e.g., outlier points that are raised off the ground plane of Mars), users will likely analyze them directly in the Track and Image Views for errors. 
Likewise, camera frustums that point in directions not consistent with their neighbors (e.g., \cref{fig:annotated}a) could also indicate issues with the initial telemetry readings or with BA, and may be further analyzed individually.
On the right-hand side of the Scene View, charts show summaries of residuals. As shown in \cref{fig:flight26}g, a histogram encodes the distribution of residual lengths across the dataset and a radial chart depicts residuals' lengths and angles. The radial chart plots each residual vector with its origin in the center of the chart, encoding each vector's end point with a low-opacity point. These charts provide an overview of residual distributions which can be used to evaluate the general efficacy of a model, before investigating particular errors. Ideal BAs result in a concentration of short residuals (i.e., a left-skewed histogram) that are evenly distributed across all angles (i.e., circular, as opposed to oblong/oblate, clouds of points in the radial chart--see \cref{fig:annotated}b). 

For \textit{Flight 26}, the visual information in the Scene View (\cref{fig:flight26}e) displays sharp turns in the flight pattern. Upon panning and tilting the 3D scene, scientists noticed a set of 3D points which should lie on the surface of Mars were computed significantly off the plane (see \cref{fig:annotated}c). %
Closer inspection reveals that these points coincide with regions where Ingenuity turned sharp corners. The short baselines in such cases have been shown to be ill-posed for 3D structure computation~\cite{Hartley04}, and BA will generally fail to adjust these.

\subsubsection{Image Grid View}
\label{sec:img-grid}
The \textbf{\textit{Image Grid View}} (\cref{fig:flight26}b) is an alternative global representation to the Scene View. This view displays all images from the dataset in a horizontal carousel, along with each image's summary charts. Each image is shown at the top of its vertical card, and is superimposed with pre- and post-BA residuals. Below the image, a histogram and radial chart  (e.g., \cref{fig:flight26}g), similar to those in the Scene View, represent the distribution of residuals in each image. At the bottom, a slope chart highlights the change in individual residual lengths from pre- to post- BA (e.g., \cref{fig:flight26}g, far right).
This view allows users to quickly visually parse through each image and identify those that might be associated with large residuals indicative of low-quality images or tracks that can be removed for more accurate stereo modeling. From this view, users can click on images of interest to visit their individual \textbf{\textit{Image Views}}.

For \textit{Flight 26}, inspection of images revealed some higher residual errors \textbf{after} BA, which is typically due to bad triangulations of specific feature tracks that skew the optimization. Post-BA values on the right-hand axis of the residual slope diagram were higher than their pre-BA counterparts on the left (see \cref{fig:annotated}d, left). Next, we entered the \textit{Image View} panel to further investigate the tracks.%

\subsubsection{Image View}
\label{sec:img-view}
The \textbf{\textit{Image View}} (\cref{fig:flight26}c) represents the residual information for an individual image and all the feature tracks to which its residuals contribute. This view can be accessed by selecting a BA camera (cone-like prism in \cref{fig:flight26}e) in the Scene View or an image card in the Image Grid View. In the Image View, each imported photo is superimposed with its corresponding residuals (e.g., \cref{fig:flight26}f). For closer inspection of the image and residuals, users can zoom using trackpad pinch-to-zoom motions or adjusting a computer mouse scroll wheel. Users can also pan via simple click-and-grab motions.  Like in the Image Grid View, a histogram and radial chart depicts an overview of the distribution of error vectors in the image. On the right-hand side of this view, every feature track that is built using the individual image in this view is shown in a scroll-able, horizontal card. On the far left of each feature track card, is a slope chart that encodes the change in error vector lengths between pre- and post-BA for the error vectors relevant to that track. 
To provide further 2D context, to the right of each slope chart sit a set of cropped images centered around the residual points that coalesce to create the track's single BA feature (e.g., \cref{fig:flight26}h). The user can click on a horizontal track card to pull up its corresponding Track View.
For \textit{Flight 26}, clicking on a specific image indeed revealed that certain feature tracks had high errors. In \cref{fig:flight26}c, the directional pattern of the bright red residual lines shows that the camera's helicopter was turning through a tight corner when it took this image. The consistency of the lines' angle and length confirmed that BA failed to minimize residuals. To check that these residuals resulted from a tracking error, not an issue with the camera models,
using the \textit{Track View}'s 3D view port (e.g., \cref{fig:flight26}e), we saw that the cameras were correctly pointed towards the ground. Finally, we were able to delete these erroneous tracks directly in the Image View panel, and re-run the stereo reconstruction process. This entire process took minutes and yielded much-improved results.

\subsubsection{Track View}
\label{sec:track-view}
Lastly, the \textbf{\textit{Track View}} (\cref{fig:flight26}d) depicts all residual information for an individual track, representing one triangulated point. The Track View can be accessed by clicking on a BA point in the Scene View or a horizontal track card on the right side of the Image View. The Track View provides granular information about each residual that contributes to a singular 3D feature. At the top of this view, each relevant residual is superimposed on top of its respective 2D image for spatial context. Below these images, a BA viewport visualizes the camera poses that contribute to each residual, along with the single point that has been triangulated. Lastly, a set of summary charts similar to those in the previous views display residuals' distribution for quick overview of track accuracy. In the Track View, users can also opt to delete points and their feature tracks if they are characterized by egregious residuals.

\subsubsection{Filtering Panel}
\label{sec:filtering}
The \textbf{\textit{Filtering Panel}} lives on the left-hand side of all of VECTOR's views. If a user alters the  parameters in one view, those parameters persist to other views. Filters only reset upon loading a new dataset or relaunching the system. Every view has a set of standard residual filters associated with type, length, angle, precision and scale. The type filter allows users to view specific residual groups, determined by the dataset. The length, angle, precision, and scale filters directly manipulate residual values shown in the view. The Scene View has additional filters for toggling rendered points or camera frustums. The Image and Image Grid View include additional sorting parameters for controlling the layout of tracks or images, respectively.

\section{Conclusions and next steps}
\label{sec:conclusions}

This paper presents the VECTOR (Visualization and Editing of Camera Tiepoints, Orientations, and Residuals) software which is currently used at NASA-JPL for evaluation and error analysis of stereo algorithms in its ground-based data processing pipeline. VECTOR was born out of a need to visualize errors and reconstruction parameters in common 3D stereo processes such as SfM and SLAM, and is being used for operations in the Mars 2020 mission.

Future work could improve VECTOR with the development of validation metrics to increase certainty about optimization. While current VECTOR users report major improvements in their ease of error finding, there is no way for users to know with certainty if errors have reached a global minimum. VECTOR was also tested on JPL datasets with hundreds of images and thousands of tracks, but scaling to datasets in the millions will need to be explored. Future work could also identify other domains that may benefit from VECTOR, within the vast amount of stereo applications~\cite{James:2012:JGR,Guidi:2004:TIP,magnani2020digital,dellepiane:2013:JCH,dewitt2000elements,grisetti2010tutorial,Matthies:2007:IJCV,maimone:2007:JFR} and other difficult optimization problems. To this end, the software is open-source for public use at \href{https://github.com/NASA-AMMOS/VECTOR}{NASA-AMMOS}.

\acknowledgments{%
The authors would like to thank the Caltech/JPL/ArtCenter Data to Discovery (D2D) Summer Data Visualization Program\cite{d2d:program, hendrie:2022:infovis} and the NASA Multi-Mission Ground Systems and Services Program (MGSS) for funding this project. This research was carried out in part at the Jet Propulsion Laboratory, California Institute of Technology, under a contract with the National Aeronautics and Space Administration (80NM0018D0004).%
}

\clearpage

\bibliographystyle{abbrv-doi-hyperref}

\bibliography{paper}

\end{document}